# Real time implementation of CTRNN and BPTT algorithm to learn on-line biped robot balance: experiments on the standing posture

Patrick Hénaff[*,1], Vincent Scesa[2], Fethi Ben Ouezdou[2], and Olivier Bruneau[2]

[1] ETIS, UMR 8051, CNRS- ENSEA-UCP, University of Cergy-Pontoise, F-9500, France
[2] LISV, University of Versailles Saint-Quentin, France

**ABSTRACT:** *This paper describes experimental results regarding the real time implementation of continuous time recurrent neural networks (CTRNN) and the dynamic back-propagation through time (BPTT) algorithm for the on-line learning control laws. Experiments are carried out to control the balance of a biped robot prototype in its standing posture. The neural controller is trained to compensate for external perturbations by controlling the torso's joint motions. Algorithms are embedded in the real time electronic unit of the robot. On-line learning implementations are presented in detail. The results on learning behavior and control performance demonstrate the strength and the efficiency of the proposed approach.*

**INDEX TERMS:** Neural control, Learning algorithms, CTRNN, Legged locomotion, Real-time systems, robotics, Biped robot

## 1 Introduction

Technological developments have enabled us to build robots with morphologies that are inspired by animals or humans. Therefore, the most recent humanoid robots are technologically complex systems, with an extremely high level of mechanical and electronic integration. They are equipped with complete perceptive systems that enable them to interact with human beings and to move in an environment built for human life. One of the most important difficulties in controlling humanoid robots is maintaining balance during walking or standing. One solution to this problem is to design controllers based on the zero moment point (ZMP) theory (Vukobratovic, 2004). Another method is to design controllers using bio inspired approaches, i.e., with some capabilities of adaptation and training, leading to the acquisition of reflexes.

[*] Corresponding Author : Patrick Hénaff, ETIS - UMR CNRS 8051, Université de Cergy-Pontoise, Site de ST Martin, 2 rue A. Chauvin, 95302 Cergy-Pontoise Cedex, France; patrick.henaff@u-cergy.fr; Tel : +33 1 34 25 28 51 ; fax :+33 1 34 25 28 39
   Email Adresses: fethi.benouezdou@uvsq.fr, bruneau@lisv.uvsq.fr.

Using biologically inspired architectures such as neural networks that are able to learn the "correct" control of the robot's equilibrium is a promising approach. For this purpose, several neural controller-based approaches have been proposed in the past. CMAC (cerebellar model articulation controllers) proposed in 1975 by James Albus (1975), are still studied in the control of legged robots. Recent studies deal with their modeling and generalization properties (Horvath & Szabo, 2007) or with their connections to other approaches like fuzzy logic (Su, Lee &Wang, 2006) or computed torque control (Lin & Chen, 2007). CMAC have been used to control the balance of biped robots (Kun & Miller, 1996) or for robust dynamic walking in simulation (Lin & Chen, 2007) and for biped robot experiments (Sabourin & Bruneau, 2005).

Recurrent neural networks (i.e., dynamic neural nets) have been extensively studied in the control of complex systems for many years (Marcua, Köppen-Seligerb & Stücher, 2008; Song & Tahk, 2001). These Artificial Neural Networks have also been used to design stable walking gaits for biped robots (Wu, Song & Yang, 2007). Several approaches were based on evolutionary synthesis (Fukuda, Komata & Arakawa, 1997), neural oscillators (Taga, Yamaguchi & Shimizu, 1991; Geng, Porr & Wörgötter, 2006), and central pattern generators (Righetti & Ijspeert, 2006; Nakanishi et al., 2004). Recently, researchers have used RNN as predictive compensator (Mizunoa, Kurodaa, Okazakib & Ohtsu, 2007) or tracking controllers with self-constructing properties. Self-constructing algorithms are very interesting approaches because they allow for optimizing on-line the neuronal controller architecture in order to insure the best control performance. In Gao & Er, 2003, a self-constructing fuzzy neural controller was proposed for the tracking control of a simulated planar robot manipulator with 2 degrees of freedom. In (Hsu, 2009) a simple growing and pruning algorithm applied to a recurrent neural network has been tested in experimentation to control one degree of freedom of a moving table with a linear ceramic motor system.

Many studies on dynamic neural controllers of robots have focused on continuous-time recurrent neural networks (CTRNN) due to their ability to be universal approximators (Beer, 2006). CTRNNs have been used for bio-inspired control because of their abilities to reproduce the full qualitative range of nerve cell phenomenology (Beer, 2006). They make it possible to show adaptivity properties based on homeostatic plastic mechanisms (Williams, 2007; Hoinville & Hénaff, 2004). Moreover, they may be a fine model for generating adaptive behavior because they can learn with the back-propagation through time algorithm (BPTT) (Pineda, 1987; Werbos, 1990; Pearlmutter, 1995; D.E. Rumelhart, Hinto & Williams, 1986; Robinson & Fallside,1987). Unfortunately, these algorithms are complex to implement in real time application, especially for the on-line training of real robots because of the shrinking gradient problem of recurrent neural networks (this gradient instability problem was first studied in Scesa, Hénaff, Ouezdou & Namoun, 2006).

There are still only a few recent in-depth studies on real time implementation, especially with the concept of learning the equilibrium reflex for a biped robot on-line. The scientific objective of the work presented in this paper is to perform an in-depth analysis of the real time performance of continuous time recurrent neural network controllers for learning balance reflexes for a biped robot. In particular, robot abilities to learn on-line with real time constraints are investigated.

Several contributions in this area have already been published (see Scesa, Mohamed, Henaff, & Ouezdou, 2005). This paper focuses on describing the real time implementation of the learning algorithms embedded into the robot control unit. Some promising new experimental results are also reported.

This paper is organized as follows. The second section deals with the learning models based on the CTRNN and BPTT algorithms. The third section presents the fundamental principles of CTRNN and BPTT. In the fourth section, the experimental biped platform called ROBIAN is described. The real time implementation of the learning algorithm based on CTRNNs and the back-propagation through time algorithm is detailed Section 5. The sixth section, describes the experimental results in two subparts: a feasibility test for the embedded learning algorithm on controlling the ROBIAN biped torso and a new approach of on-line learning of the equilibrium reflex. Finally, the last section presents the conclusions and potential further developments stemming from this work.

## 2   Learning equilibrium with CTRNN

It is well known that the human torso attempts to stabilize the whole body when walking (Setiawan, Hyon, Yamaguchi & Takanishi, 1999; Kubica, Wang, & Winter, 2001; Hyon, 2009). It is also known that biped robot balance can be considered to be a global behavior because the upper part (the torso) and the lower part (the legs) of the robot interact with and disturb each other (see Figure 1) during walking. This internal interaction can be taken into account in the synthesis of the biped robot controller. For example, J Morimoto et al. (2006) show that a biped robot can stop and walk using simple sinusoidal desired trajectories with their phases adjusted by a coupled oscillator model.

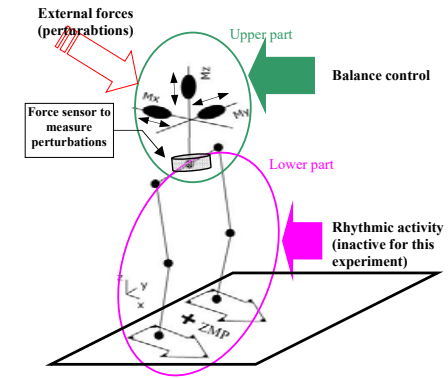

Fig. 1.  Control of the biped robot's balance: perturbations caused by unknown external forces can be measured with a force sensor placed between upper and lower part.

During the walking and halting phases, the robot's balance can be affected by perturbations like external forces applied to its body, including the upper part. One way to control the balance of the robot is to drive the movements of the upper part (i.e., the torso) in order to minimize the perturbations (forces and moments) exerted on the lower part of the robot.

To measure this perturbation, the robot is equipped with a six-force sensor fixed between the torso and the lower part (see Figure 1). Then, if the forces and moments measured by the sensor are close to zero, the equilibrium of the robot is assumed to be controlled (Mohammed, Gravez & Ouezdou, 2004).

To achieve this goal, the purpose of this experiment is to learn on-line, using a CTRNN, how to control the robot torso in order to reduce the external perturbations measured by the sensor. This learning is carried out with the biped robot in the standing posture when unknown external forces are applied to its torso.

The control of robot balance must take into account the real features of the mechanism and the external phenomena that are not modeled in the simulation (friction, motor properties, attrition, ground slope and passive prosthesis use…). Information on these phenomena is often only available through their effects on the total energy of the system. Hence, taking into account these phenomena into the equilibrium control is a difficult task. To avoid this obstacle, the controller should be able to adapt its behavior in real time, following a cost function that incorporates information on these phenomena. Moreover, in an optimal control approach, time variations must be included in the adaptation process.

To meet these conditions, dynamic recurrent neural networks with back-propagation through a time learning process was chosen as the most appropriate network. The proposed learning control architecture is shown in Figure 2. The outputs of the network are positions $(X, Y, Z)$ that the torso mechanism then has to reach. Thus, at each time step, the net modifies the actuator speed and accelerations are generated. Consequently, forces between the upper and lower parts of the robot will be produced to compensate for the external perturbations.

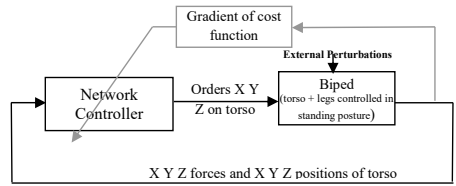

Fig. 2. Principle of the on-line learning balance control scheme. Grey lines represent the learning process with cost computation and modification of the net. Each joint position is measured with an incremental electronic sensor. Forces between the upper and lower parts are measured with the six-force sensor. All signals are fed back to the inputs of the control boards of the electronic controller unit (see Section 5).

The inputs should show the network the current state of the system. They must be enough representatives to compute a correct control response. They consist of the measured components provided by the 6-forces sensor and the positions of the current motors given by incremental encoders.

To carry out the on-line learning of an optimal control, the parameters of the net must be adapted in real time while the system is running. Back-propagation through time (BPTT), with its ability to integrate the error in the network at each instant is an appropriate solution to solve the parameter adaptation problem. In BPTT, the network is first unrolled in time creating a layer per time step. Then, this algorithm back-propagates the output error on these virtual temporal layers as the classical back-propagation algorithm does on existing ones. The result is the computation of the error gradient in the network and its integration in time. Then, the parameters are modified by the gradient descent algorithm, and the response of the net approaches the desired behavior. Instead of the output error, a cost function that represents the aim of the control can be used for the learning stage. The aim is to bring the forces measured by the sensor closer to zero so that the cost is define by the sum of the squared force components measured by the six axes force sensor in the $X, Y$ and $Z$ directions:

$$Cost(t) = F_X^2(t) + F_Y^2(t) + F_Z^2(t) \qquad (1)$$

This cost function is given for the learning algorithm through a gradient computation for each net output:

$$e_j(t) = \frac{\partial Cost(t)}{\partial output_j} \qquad (2)$$

This equation expresses the influence of each command on the cost function at time $t$. BPTT will use this gradient to minimize the integral of the cost function (Equation (1)) over the learning time window. Thus, the cost function computed to optimize the control has the following expression:

$$E = \int_{t_0}^{t} [F_X^2(t) + F_Y^2(t) + F_Z^2(t)] dt \qquad (3)$$

Instead of computing the exact expression of (3), which would require a strict knowledge of the system, it is possible to use a signal that carries the same information. For the torso experiment, it is defined by the following equations:

$$e_Z(t) = F_Z(t) \quad e_X(t) = F_X(t) \quad e_Y(t) = F_Y(t) \qquad (4)$$

## 3    Principles of CTRNN and BPTT

Neurons of a dynamic recurrent neural network are driven by the following equation, derived from (Beer, 2006; Pineda, 1987; Pearlmutter, 1995):

$$T_i \frac{\partial y_i}{\partial t} = -y_i + f(x_i) \qquad (5)$$

where

$y_i =$ output of neuron $i$

$x_i =$ weighted sum of the $j^{th}$ neuron inputs

$f =$ activation function ($tanh$ or sigmoid)

$T_i =$ time constant of the $i^{th}$ neuron

### 3.1    Propagation algorithm

The input data provided to the network are propagated to generate output responses. The propagation depends on the intrinsic parameters of the network and the neurons, which are given as follows for the $j^{th}$ neuron:

$\omega_{ij} =$ weight of the connection from neuron $i$ to neuron $j$

$b_j =$ bias of the $j^{th}$ neuron

$T_j =$ time constant of the $j^{th}$ neuron.

By the progressive (i.e., explicit) Euler approach, for the $\Delta t$ time step, and the discrete propagation equation depends on the following relations (Nguyen & Cottrell, 1997):

- Input sum for the $j^{th}$ neuron :

$$x_j(t) = \sum_i [\omega_{ij} \cdot y_i(t - \Delta t)] + b_j(t) \qquad (6)$$

- Output of the $j^{th}$ neuron :

$$y_j(t + \Delta t) = \frac{\Delta t}{T_j} \cdot f(x_j(t)) + \left(1 - \frac{\Delta t}{T_j}\right) \cdot y_j(t) \qquad (7)$$

In this equation, the $\frac{\Delta t}{T_j}$ term is a scale parameter. Its value, which cannot be equal to zero, lies in the interval $]0{:}1]$ and expresses the response speed of the $j^{th}$ neuron.

By defining a new variable $S_j = \frac{\Delta t}{T_j}$, Equation (7) can be rewritten as follows:

$$y_j(t + \Delta t) = S_j \cdot f(x_j(t)) + (1 - S_j) \cdot y_j(t) \qquad (8)$$

Fig. 3 graphically represents the principles of the propagation through time algorithm. At time $t$, each neuron (except input neurons) receives the $(t - 1)$ outputs of each other neuron.

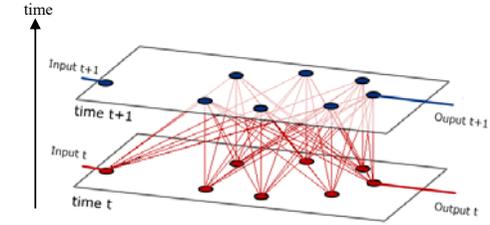

Fig. 3. Equivalent scheme of propagation in time. Example of a 1-6-1 network (1 input, 6 hidden neurons, 1 output) on two time steps.

### 3.2    Truncated Dynamic BPTT learning algorithm

The time constant parameter of each unit reflects the dynamic aspect of the net. However, the learning process must be able to teach it. For this purpose, Pearlmutter (1995) proposed, a dynamical BPTT version for which the time constant, the weights and the bias parameters can be adjusted. The only remaining problem is the memory needed for BPTT algorithms. To compute the exact error gradient, the algorithm has to store all of the network states from the beginning to the current time step. To avoid this excessive memory use, a truncated version, in which the states of the net are only stored during a time window that follows the current instant, can be used (Williams & Zipser, 1989; Williams & Peng, 1990). Therefore, it only computes an approximation that approaches the exact gradient as the length of the time window increases. Therefore, the learning algorithm used in the experiments will be a truncated dynamic version of the back-propagation through time algorithm, called TDBPTT.

This learning process begins with an error computation at the output of the net. It represents the remaining gap between the current response and the aim of the learning. This error could simply be a squared difference between the desired and computed response. It could also be the result of a more complex computation, e.g., the equilibrium of a biped robot. The objective of the learning process consists of modifying the network parameters (weights, biases and time scale) in order to minimize a desired criterion. In a control application, this criterion would be the gap between the desired state of the system and the actual state. In an identification process, the criterion would be the error between the neural model and the system being taught.

At first, the algorithm computes an error function that corresponds to the criterion to be minimized. This function, $E$, is the error integral between $t_0$ and the current time step, $t$, expressed as follows:

$$E = \int_{t_0}^{t} \sum_{j=1}^{n_s} e_j(\tau).d\tau \quad (9)$$

where $n_s$ is the number of output neurons, $e_j(\tau)$ is the output error of the $j^{th}$ neuron stored at time $\tau$ and $t_0$ is the beginning of the integration window. The BPTT algorithm carries out gradient based learning. Thus, the parameters are modified by the negation of the error gradient of each parameter:

$$\Delta \omega_{jk} = -\eta \frac{\partial E}{\partial \omega_{jk}} \quad \Delta b_j = -\eta \frac{\partial E}{\partial b_j} \quad \Delta T_j = -\eta \frac{\partial E}{\partial T_j} \quad (10)$$

where $\eta$ is the learning step. The parameters of the learning process are the learning step and the time window width $(t - t_0)$.

*Calculation of the gradient components*

The back-propagated cost attached to the $j^{th}$ neuron is defined by

$$Z_j(\tau) = \frac{\partial E}{\partial y_j(\tau)} \quad (11)$$

and $Z_k$ is the back-propagated cost attached to the downstream neuron $k$ for each components of the gradient.

The calculation of the gradient components for each parameter is carried out on the network "unrolled in time". Each component is estimated for the completely unrolled network, but it results in the values of instantaneous components of the gradients, called the back-propagated gradient components. The total value is the integral, over the unrolled network, of the back-propagated values given by Equation (14), for the $j^{th}$ neuron connected to the $k^{th}$ neuron by the weight $\omega_{jk}$.

By the formula of derivative of the composed functions, the back-propagated components of the gradient can be expressed as follows (see Appendix A for details):

- Gradients on weights:

$$\frac{\partial E}{\partial \omega_{jk}} = \int_{t_0}^{t} \left[\frac{\partial E}{\partial \omega_{jk}}(\tau)\right].d\tau = S_k. \int_{t_0}^{t} [Z_k(\tau).f'(x_k(\tau - \Delta t)).y_j(\tau - \Delta t)].d\tau \quad (12)$$

where $f'$ is the derivative of the sigmoidal activation function.

- Gradients on bias:

$$\frac{\partial E}{\partial b_j} = \int_{t_0}^{t} \left[\frac{\partial E}{\partial b_j}(\tau)\right].d\tau = S_j. \int_{t_0}^{t} [Z_j(\tau).f'(x_j(\tau - \Delta t))].d\tau \quad (13)$$

- Gradients on scale parameters:

$$\frac{\partial E}{\partial T_j} = \int_{t_0}^{t} \left[\frac{\partial E}{\partial T_j}(\tau)\right].d\tau = \frac{-S_j^2}{\Delta t}. \int_{t_0}^{t} [Z_j(\tau).(f(x_j(\tau - \Delta t)) - y_j(\tau - \Delta t))].d\tau \quad (14)$$

*Calculation of back-propagated costs:*

The value of the back-propagated cost $Z_j$ expresses the influence of the output of the $j^{th}$ neuron, at a given time on the current cost. Its computation uses the instantaneous error values $e_j(\tau)$, calculated outside the network by the comparison of the current state of the network and the objective to be reached.

For each neuron at one instant, $\tau$ of the past, the back-propagation cost comes from three sources. Indeed, the back-propagation "goes up" step-by-step through the neurons from the current activities of the network to the oldest states. The back-propagated costs $Z_j^1(\tau)$ and $Z_j^2(\tau)$ are present if the neuron is static or dynamic, and $Z_j^3(\tau)$ appears only for dynamic neurons. $Z_j^1(\tau)$ corresponds to the error coming at the following instants from the synapses connecting the neurons. $Z_j^2(\tau)$ represents the error coming directly from the output neurons (it does not exist if the neuron is hidden). $Z_j^3(\tau)$ corresponds to the error coming at the following instants by means of the internal dynamics of the neuron itself. Hence, the total backpropagation cost $Z_j(\tau)$ is

$$Z_j(\tau) = Z_j^1(\tau) + Z_j^2(\tau) + Z_j^3(\tau) \qquad (15)$$

Figure 4 represents the calculation of these three elements of the backpropagated cost through the $j^{th}$ neuron.

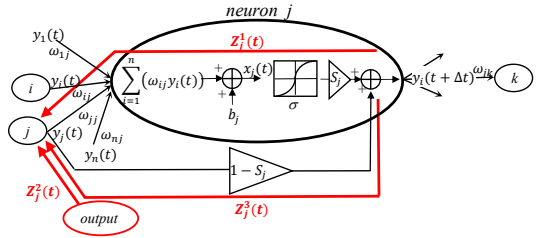

Fig. 4. Ways followed by the three elements of the backpropagated cost through a neuron.

The calculation of these elements is detailed in Appendix B, and it finally yields the total back-propagated cost:

$$Z_j(\tau) = \sum_{k=1}^{n} [Z_k(\tau + \Delta t). S_k. f'(x_k(\tau)). \omega_{jk}(\tau)] + \frac{\partial e_j(\tau)}{\partial y_j(\tau)}. \Delta t + Z_j(\tau + \Delta t). (1 - S_j) \qquad (16)$$

The integral, present in the total cost expressed by the Equation (9), which made it possible to generate an optimal controller, is consequently distributed over time and over the output neurons. Nevertheless, the calculation of the components of the gradient, reintroduces this integrating aspect, thus guaranteeing the convergence towards an optimal solution over time. This explains the presence of the multiplying term, $\Delta t$, in the expression $Z_j(\tau)$ that takes into account the temporal integral of the errors coming from outside.

## 4   ROBIAN biped

The learning approach is validated on the ROBIAN biped prototype (Konno, Sellaouti, Amar & Ouezdou, 2002). ROBIAN consists of two different parts: a locomotion system (lower limbs, i.e, legs) and a special torso mechanism (upper part) that is described in the next section. Each leg has a total of seven dofs (Sellaouti, & Ouezdou, 2005): three actuated dofs for the hip, one actuated for the knee, two actuated for the ankle and one passive for the foot, providing a flexible foot system (see Figure 5).

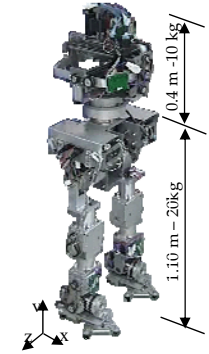

Fig. 5. The ROBIAN biped: total height is 1.50 m and weight is 30 kg.

### 4.1   A simplified torso for balance control

The human upper body can be modeled with 13 degrees of freedom, which is not an easy to mechanism control. In order to identify a minimal mechanism of the human upper body that can mimic it when walking, a previous analysis of the six wrench components exerted by the upper part of a virtual manikin on the locomotion apparatus was conducted. It led to the identification of two coupling relations between the upper and lower body. Then, the dynamic equivalence between mechanisms allowed us to identify a trunk mechanism with 4 degrees of freedom that is able to reproduce the dynamic effects of the upper limbs during the walking gait (to understand in detail the advantage of using a simplified torso for balance control, see Mohammed, Gravez & Ouezdou, 2004).

ROBIAN's trunk (see figure 6 and 7) is an original mechanism with 4 dofs: one rotational (R) and three translational (P) movements. It is important to note that the three elements, $C_2$, $C_3$ and $C_4$, are directly connected to element $C_1$ by prismatic joints (P) and the whole system is connected by a rotational joint (R) to element $C_0$.

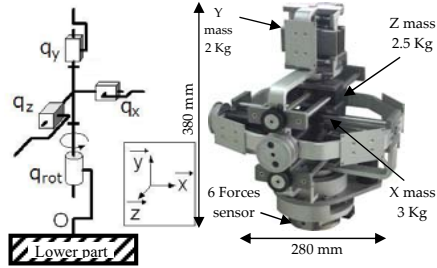

Fig. 6. The ROBIAN trunk is an RPPP mechanism. On the picture, one can see Y and Z masses that follow their prismatic joint (the X mass is hidden). Total weight is 13 kg.

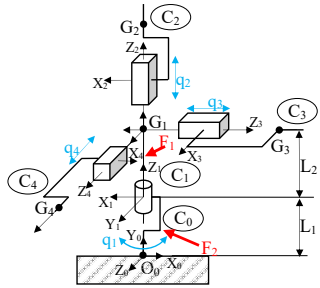

Fig. 7. Modeling of trunk R3P subjected to external forces, F1 and F2

The sagittal plane of the trunk R3P is $(O_0, X_0, Y_0)$, and the frontal plane is $(O_0, Y_0, Z_0)$. The properties of the R3P mechanism are presented in Table I. The $C_i$ (i=1.., 4) elements represent the mobile masses that are used to compensate for the external disturbances.

The simulation results obtained in Mohamed, Gravez, Bruneau & Ouezdou, 2002 demonstrate the existence of coupling relations between torques and forces exerted by the legs on the trunk during walking:

$$M_X = k_1 . F_Z \quad \text{and} \quad M_Z = k_2 . F_X \quad (17)$$

where $F_X$, and $F_Z$ are the components of the measured efforts exerted by the support to the trunk at point $O_0$.

The efforts measured to the trunk by the sensor on point $O_0$ are given by (see Zaoui, Bruneau, Ouezdou &. Maalej, 2009 for more details):

$$\begin{cases} \vec{F_0} = \sum_{i=0}^{4} m_i(\vec{\gamma}_{G_i} - \vec{g}) - \sum_{i=1}^{2} \vec{F_i} \\ \vec{M_0} = \sum_{i=0}^{4} \left( \vec{h}(i, O_0) - \vec{M}(m_i \vec{g}, O_0) \right) - \sum_{i=1}^{2} \vec{M}(\vec{F_i}, O_0) \end{cases} \quad (18)$$

where

- $\vec{h}(i, O_0)$: Derivation of element $C_i$ angular momentum calculated at point $O_0$
- $\left\{ \frac{\vec{F_0}}{\vec{M_0}} \right\}$: Efforts exerted by the frame on $C_0$ at point $O_0$
- $\vec{F_1}$ and $\vec{F_2}$: External forces applied respectively to $C_1$ and $C_0$ ($C_1$ and $C_0$ are the elements in contact with the external environment of the robot) with $F_1 = [F_{1X}, F_{1Y}, F_{1Z}]^T$ and $F_2 = [F_{2X}, F_{2Y}, F_{2Z}]^T$
- $\vec{\gamma}_{G_i}$: Acceleration of $C_i$ elements.

Note that if body $C_0$ is fixed, then $\vec{\gamma}_{G_0} = \vec{0}$ and $\vec{h}(0/0, O_0) = \vec{0}$.

Furthermore, because $G_0 = O_0$: $\vec{M}(m_0 \vec{g}, O_0) = \vec{0}$.

TABLE I.

MASS AND INERTIA PROPERTIES OF THE ROBIAN TORSO

| | | |
|---|---|---|
| $C_0$ | $m_0$ (kg) | 0.5 |
| $C_1$ | $m_1$ (kg) | 5 |
| $C_2$ | $m_2$ (kg) | 2 |
| $C_3$ | $m_3$ (kg) | 3 |
| $C_4$ | $m_4$ (kg) | 2.5 |
| Total inertia | Iy (kg.m²) | 0.28 |
| $L_1$ | (m) | 0.1 |
| $L_2$ | (m) | 0.2 |

These dynamic equations show that the movements of the three masses can compensate for unknown external forces applied to the trunk. Thus, in using BPTT to learn this compensation, the cost function and the gradient components for the learning stage will use these equations.

# 5 Real time implementation

The real time implementation of the neural controller is a necessary objective to conduct an efficient learning control. The general architecture of the robotic application is based on the exchange of information between different systems: a PC user, an industrial electronic control unit and the ROBIAN robot (Fig. 8.).

The electronic unit is built with one server board (SH3 RISC, 96 MHz, 16 Mb of RAM, 128 kb of flash memory, 32 kb of Shared memory) and 8 client boards (SH3 RISC, 96 MHz, 8 Mb for RAM, 128 kb for the flash memory). These 8 boards are dedicated to the control of the 16 DC motors and to the measurement acquisition (2 DC motors per board). The user computer is dedicated to supervising the robot. It communicates with the server board via an Ethernet protocol, which exchanges data with the eight client boards via a special local area network. These client boards exchange with the robot through point-to-point wiring. Programs are run on the server board and are built around the use of objects representing the actuators, the sensors of the robot and their parameters. These objects are written in the shared memory of the electronic unit. The control of each motor is regulated locally via the control boards.

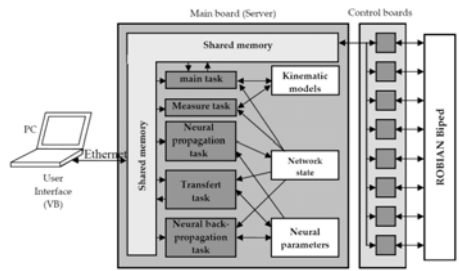

Fig 8. Control architecture of the ROBIAN platform

## 5.1 Multi task organization

The neural networks algorithms, written in the C language, are implemented in the server board. A user interface, written in the Visual Basic graphic language, gives access to all the parameters of the algorithms, to the measurements taken on the robot and to the neural network parameters. The entire application is divided into 5 specific tasks, as shown in Figure 8: the "Main" task, which focuses on most of the information exchanged and manages the objects of the shared memory, the "Measure" task, the "Transfer" task, the "Propagation" task and the "Back-propagation" task.

## 5.2 Scheduling of learning tasks

*Temporal evaluation*

Other experiments have shown that the fastest oscillation period that can be applied to the biped robot in its frontal plan is 500 ms. Thus, the computing time required by the ROBIAN control must be limited to 10 ms in order to maintain sufficient control.

However, the mathematical complexities (number of implied operations) associated with the propagation functions and the training of the neural controller are $O(n^2)$ and $O(W_T n^2)$, respectively, with $n$ being the number of neurons and $W_T$, the width of the time window used. Consequently, if the two functions are computed sequentially, then the number of operations will evolve out of $O(n^2 + W_T n^2)$, and the associated computing time will be able to exceed the limit required. For this reason, the propagation and the backpropagation are separated into two dependent and synchronized tasks. Then, according to the ratio of the computing time (related to the width of the time window) of each task for one training, several propagations in the network are carried out.

Table II compares the time performance of propagation and backpropagation. The sampling of the propagation task depends on the computing time necessary for the propagation in the network and, consequently, on the number of parameters defined with the network structure. For example, the propagation time is 2 ms through a completely recurring network consisting of 10 neurons (and 3 ms for 14 neurons and time window of 10). The back-propagation, which is computed along the time window, has an execution time that depends on the width of the time window. For a 10 neuron network with a time window memorizing the 20 preceding states, the back-propagation task will require 50 ms (60 ms for 14 neurons and a time window of 10).

TABLE II.

TIME PERFORMANCES OF PROPAGATION AND BACKPROPAGATION

|  | Propagation | Backpropagation |
|---|---|---|
| Complexity | $O(n^2)$ | $O(T_W \cdot n^2)$ |
| Computational Time($n$=10,$W_T$=20) | 2 ms | 50 ms |
| Computational Time($n$=14,$W_T$=10) | 3 ms | 60 ms |

The algorithmic complexity of the learning tasks is calculated involved in the training, where $n$ is the number of neurons in the net, $W_T$ is the Time Window width (number of net states stored). 14 neurons and $W_T$ =10 are setup values for the experiments described in Section 5.

Thus, even taking into account the commutation time between the tasks, these two sampling periods remain approximately equal to 2 ms and 50 ms (see next subsection). Figure 9 is a snapshot taken by the numerical oscilloscope of the user interface, which is connected to the electronic control unit. It shows the dependence and synchronization between the propagation and back-propagation tasks.

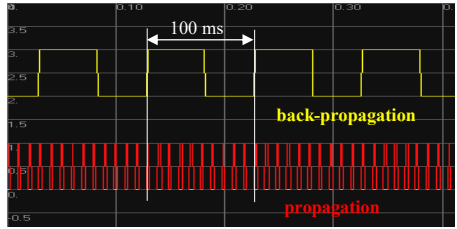

Fig. 9: Visualization of activations of propagation and backpropagation tasks computed in real time by the 10 neurons and there is one backpropagation for 10 propagation step states.

The learning procedure based on the BPTT algorithm uses neuron outputs to calculate backpropagated costs and gradients. However, these same values are modified by the propagation procedure carried out in parallel. It is then not possible to use the same set of variables for the two tasks. Thus, two different tables have been set up in the neuronal algorithms to memorize the states of the network.

*Synchronization*

For each new iteration of learning, it is necessary to transfer the states from the network generated by the propagation task to the learning task. This procedure, which consists of copying one table into another, is placed upstream of the back-propagation. It is called "temporal photograph" because it memorizes the current instantaneous state of the network. Thus, this table contains the previous states of $W_T$ that are present in the temporal window.

The learning procedure gathers four principal stages: temporal photograph (A), backpropagated costs calculation (B), gradient components computation (C), and parameter modification (D). Stages (B) and (D) are based on the values memorized by the temporal photograph and do not interact with those used by the propagation. On the other hand, stages (A) and (D) copy the network states and modify the parameters. Therefore, their executions are directly related to that of the propagation, as shown in Figure 9. The stage (A) must start only when the propagation (P) is finished. Indeed, it is not necessary for the values that make up a state of the network to be modified when copying this state. To avoid this problem, stage (A) initially checks that the propagation is not running. If the propagation is running, then stage (A) waits until it ends before carrying out the copy. The end of the propagation is announced by a binary semaphore, with a value of 0 (propagation stopped) or 1 (propagation in progress).

In the same way, stage (D) must be carried out when the propagation is stopped. Indeed, this stage modifies the parameters of the network. Thus, it is essential that these parameters are not used when their values are updated. Therefore, this stage (D) waits until the next propagation is finished by observing the semaphore value. If the propagation is not running when launching task (D), it will wait until the propagation launches (propagation in progress) and then finishes (propagation stopped). This will ensure that the propagation will not start during the modification of the parameters. The computation times for this example are given by the table in Figure 10. The duration does not include the waiting periods due to the synchronization process.

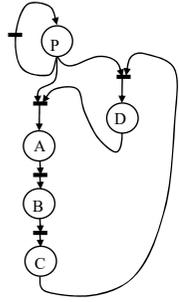

| Task | Duration |
|---|---|
| Propagation (P) | 2 ms |
| temporal Photograph (A) | 4 ms |
| $Zj(\tau)$ Computation (B) | 19 ms |
| gradient Computation (C) | 21 ms |
| Parameters Modification (D) | 1 ms |

Figure 10: On the left, description of synchronization between the propagation task and the backpropagation procedures. On the right, computation times of the different neural procedures for a 10-neuron network.

Figure 11 details the execution times of the propagation task, and the other tasks necessary for the learning process. The duration of the induced synchronization waits are also included. In this example, the network consists of 10 neurons and the temporal window has a width of 20 memorized states. The propagation period task is 10 ms, whilst that of the learning task is 100 ms, and the sampling rate of the measurement is 1 ms. Stage (A') corresponds to the 3 ms wait required before the beginning of the temporal photograph (A). The stage (D') represents waiting for 5ms before modifying the network parameters. Note that this stage waits until the propagation starts and finishes before allowing the modification of the parameters (D).

These details clearly show the loop-time for the control. The time necessary to refresh only the commands of the DC motors, depending on the propagation, is about 2 ms. Taking into account the learning aspects, the loop-time for the control (i.e., the time required to refresh the commands of the DC motors and the network controller) is 53 ms. Hence, it is sufficiently compared to the whole dynamics of the robot, which has an equivalent time constant of 1.5 seconds. Thus, the loop time does not degrade the robot system and the performance of the control laws.

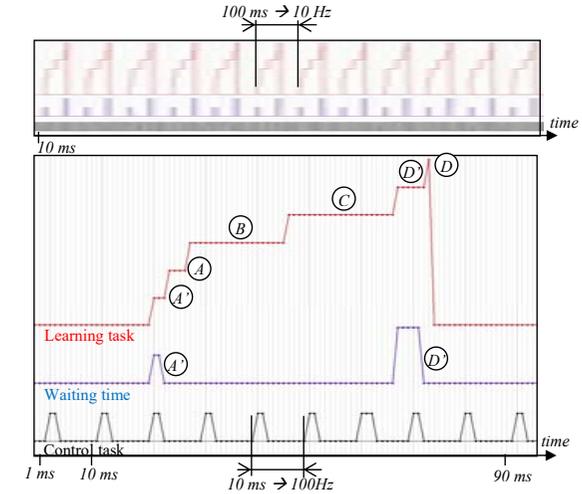

Fig. 11. Synchronization of propagation and learning extracted from the user interface: general view (up) and zoom (bottom) for the 10-neuron network. The duration of the propagation task is 3 ms and its period is 10 ms. The total duration of the back-propagation task is 53 ms and its period is 100ms.

## 6 Experimental Results

### 6.1 Settings for experiments

For this experiment, a 6-5-3 network is used: 6 inputs, 5 hidden neurons and 3 outputs (14 neurons). The number of hidden neurons was selected after some testing. At the beginning of the learning, the weight values of the net are randomly chosen. The time constants are selected from a wide enough range that the net behavior is stable, i.e., that there is no oscillation of the control.

TABLE III.

NUMERICAL VALUES OF LEARNING ALGORITHM PARAMETERS

| parameter | name | Initial value |
|---|---|---|
| Number of neurons | $n$ | 14 |
| Initial weights | $\omega$ | Randomly chosen in range of [-5;5] |
| Width of time window | $W_T$ | 10 steps |
| Time constant values | $T$ | Randomly chosen in range [0s;1s] |
| Learning step | $\eta$ | 0.001s |

*6.2 Learning the balance control with Dynamic BPTT*

For this approach to the robot's equilibrium control, the experiment is only based on the measured wrench between the upper part and the locomotion system given by the 6-component force sensor. The neural torso controller aims to keep these forces and moments close to zero for the X and Z axes and to weight the value for the Y-axis, even when an external perturbation is applied to the torso. The controller only tries to compensate for external perturbations generated manually on the torso to emulate the coupling effects in a walking gait. The experiment consists of following stages:

- ROBIAN stays in the standing posture with all its joints controlled.
- Two successive stages are carried out: learning for 60 seconds and then steering.
- During the learning stage (successive short periods), unknown external forces are manually applied to the torso in X and Z directions (for more comprehension, see Fig. 12, Fig. 13).
- The rotational motion of the torso is locked for this experiment.
- The $X$ and $Z$ moment components are not required for the net computation because their effects can be deduced from the component forces due to the coupling relations (Equations (16)). Moreover, as the rotational motion is locked, $M_Y$ data are also unnecessary.

The learning stage consists of applying forces to each axis, successively. Indeed, each mass is able to compensate only for the forces that appear on its axis. The amplitudes of the applied perturbation and their periods are voluntarily variable in order to show the network different kinds of situations. The values of these external forces are in the maximal range corresponding to the limit of the biped toppling.

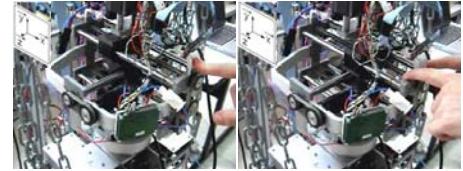

Fig.12. External forces are applied manually during successive short moments following the X direction here (left: pushing; right pulling).

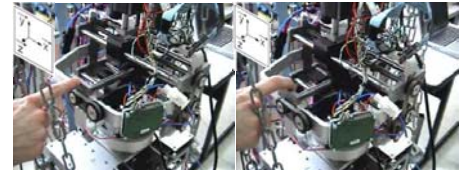

Fig.13. External forces are applied manually during successive short moments following the Z direction here (left: pushing, right: pulling).

*6.3 On-line learning of disturbance compensation*

The aim of the learning stage is to learn to compensate for unknown external disturbances. The time evolutions of responses during the learning stage are depicted in Figures 14, 15, and 16. The total duration of the learning stage is 120 seconds. This is the time required by the neural controller to learn the decoupling between the axes when external forces are applied to the ROBIAN torso. The learning algorithm starts at t=4.5s. Until t=11.4s, no external forces are applied, so the measured forces are close to zero ($F_Y$ close to the ROBIAN torso weight value), and the net does not modify its parameters. Next, during the X-axis perturbation period, the net modifies its X output command to generate a stronger and faster response to the external forces until it becomes insensitive to the disturbance (around t=20s). Then (from t=18 s to 22 s), the Z-axis becomes sensitive to the X-perturbation, and the neural controller explores its control space and learns the decoupling between the two axes. Next (after t=34s), on each axis, the mass has to move in the same direction as the force.

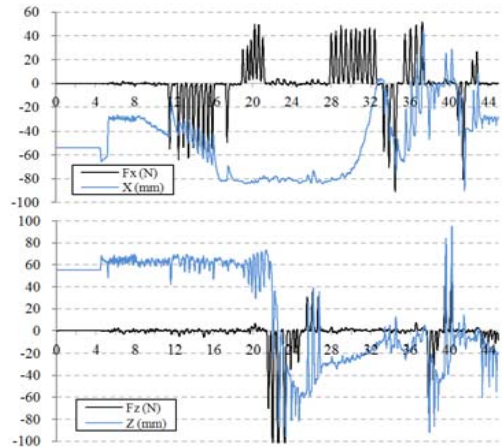

Fig.14. Beginning of on-line learning for the X-axis and Z-axis during time (s). The learning algorithm starts at t=4.5s. The neural controller tries to learn the decoupling between the two axes.

Next, from t=42s (Fig. 14) to t=63s (Fig. 15), it progressively modifies this coupling by decreasing the influence of the X force on its Z control law. From t=62 s, it is clear from Fig. 15 that the network can compensate for each pulse of perturbation due to lateral (Z axis) or frontal (X axis) pushes against the robot. After this convergence period, the learning continued for about 40 seconds. During that time, the previous changes were carried out (with forces in the opposite direction and more switches between X and Z external perturbation forces). At the end, the network emulated a controller that was able to compensate for a portion of the external forces applied (see Fig. 16). The learning algorithm stops at t=120 s; when a perturbation occurs on an axis, it does not compensate. As far as the Y-axis is concerned, the learning process manages to find motions to compensate for external Y forces in the same way as it does for the two other axes. There was no coupling between this axis and the others.

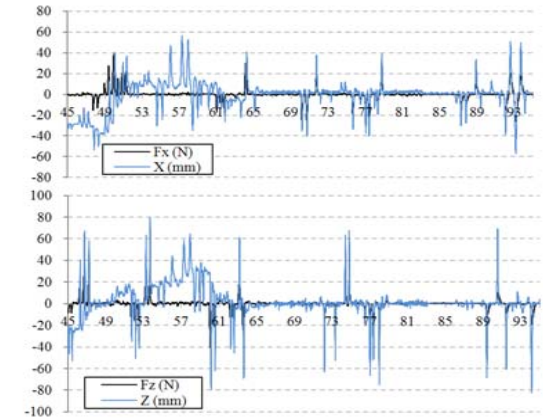

Fig. 15. Middle of on-line learning for the X-axis and Z-axis during time (s). The neural controller learns the decoupling between the two axes, and then compensates each perturbation.

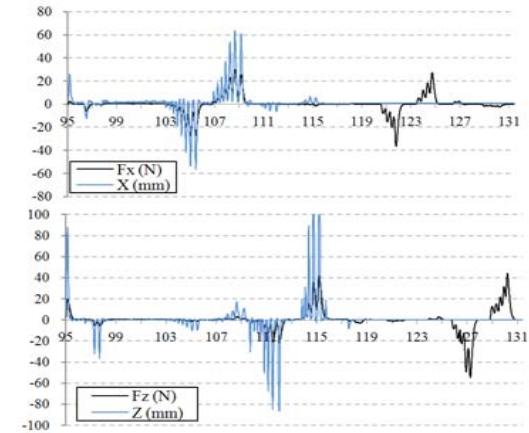

Fig. 16. End of on-line learning for the X-axis and Z-axis during time (s). Until t=120s, the neural controller learned the decoupling between the two axis. After, the learning algorithm stops at t=120s, when a perturbation occurs on an axis, it is not compensate.

## 6.4 External disturbance compensation after learning

Once the learning stage is finished, the net's parameters (biases, weights and time constants) are fixed and stored. The steering consists on using them without adding any new modification.

During the neural control period, the same forces are applied, but the 6-force sensor also measures the influence of the motion of the mass. The difference between the force shapes in the two periods represents the compensation brought about by the mass motion. In a previous study (Scesa, Mohamed, Henaff & Ouezdou, 2005), it was shown that neural control brings a decrease of about 50% in the measured forces on each axis.

## 7 Conclusion

This paper addresses the real time implementation of on-line learning control with recurrent Neural Networks. The aim of the control is to compensate for external perturbations due to lateral or frontal pushes against the robot. An experiment is carried out to control the balance of a biped robot in its standing posture. CTRNN neural nets and BPTT learning have been used to optimize a cost function which explains the rejection of perturbations. The algorithm is embedded in the real time electronic unit of the robot, and the on-line learning implementation is very detailed. Preliminary results on the learning behavior and the control performances are presented. The implementation enables to focus on the real performance of the neural network controller algorithms and shows that CTRNN are able to learn the balance reflex on-line, when controlling the robot. These experimental results show the strength and efficiency of the proposed approach.

ACKNOWLEDGMENT

The authors would like to thank the French BIA Company, which provides the real time electronic unit that controls ROBIAN and funded the PhD thesis of V. Scesa via a CIFRE contract.

## APPENDIX A

*Details of gradient components calculation*

The backpropagated cost attached to the $j^{th}$ neuron is defined by:

$$Z_j(\tau) = \frac{\partial E}{\partial y_j(\tau)} \quad (A1)$$

and $Z_k$ the back-propagated cost attached to the neuron downstream $k$, then each components of the gradient can be written as :

- **Gradients on weights**

$$\frac{\partial E}{\partial \omega_{jk}} = \int_{t_0}^{t} \left[ \frac{\partial E}{\partial \omega_{jk}}(\tau) \right] . d\tau$$

$$= \int_{t_0}^{t} \left[ \frac{\partial E}{\partial y_k(\tau)} \frac{\partial y_k(\tau)}{\partial \omega_{jk}(\tau)} \right] . d\tau$$

$$= \int_{t_0}^{t} \left[ Z_k(\tau) \frac{\partial y_k(\tau)}{\partial \omega_{jk}(\tau)} \right] . d\tau$$

$$= \int_{t_0}^{t} \left[ Z_k(\tau) \frac{\partial y_k(\tau)}{\partial x_k(\tau)} \frac{\partial x_k(\tau)}{\partial \omega_{jk}(\tau)} \right] . d\tau$$

$$= S_k . \int_{t_0}^{t} \left[ Z_k(\tau) . f'(x_k(\tau - \Delta t)) . y_j(\tau - \Delta t) \right] . d\tau \quad (A2)$$

$f'$ is derivative from the sigmoid activation function

- **Gradients on bias**

$$\frac{\partial E}{\partial b_j} = \int_{t_0}^{t} \left[ \frac{\partial E}{\partial b_j}(\tau) \right] . d\tau$$

$$= \int_{t_0}^{t} \left[ \frac{\partial E}{\partial y_j(\tau)} \frac{\partial y_j(\tau)}{\partial b_j(\tau)} \right] . d\tau$$

$$= \int_{t_0}^{t} \left[ Z_j(\tau) \frac{\partial y_j(\tau)}{\partial b_j(\tau)} \right] . d\tau$$

$$= \int_{t_0}^{t} \left[ Z_j(\tau) \frac{\partial y_j(\tau)}{\partial x_j(\tau)} \frac{\partial x_j(\tau)}{\partial b_j(\tau)} \right] . d\tau$$

$$= S_j \cdot \int_{t_0}^{t} \left[ Z_j(\tau) \cdot f'(x_j(\tau - \Delta t)) \right] d\tau \quad (A3)$$

- **Gradients on scale parameters**

$$\frac{\partial E}{\partial T_j} = \int_{t_0}^{t} \left[ \frac{\partial E}{\partial T_j}(\tau) \right] d\tau$$

$$= \int_{t_0}^{t} \left[ \frac{\partial E}{\partial y_j(\tau)} \frac{\partial y_j(\tau)}{\partial T_j(\tau)} \right] d\tau$$

$$= \int_{t_0}^{t} \left[ Z_j(\tau) \frac{\partial y_j(\tau)}{\partial T_j(\tau)} \right] d\tau$$

$$= \frac{-S_j^2}{\Delta t} \cdot \int_{t_0}^{t} \left[ Z_j(\tau) \cdot \left( f(x_j(\tau - \Delta t)) - y_j(\tau - \Delta t) \right) \right] d\tau \quad (A4)$$

APPENDIX B

*Backpropagation in a dynamic neuron.*

- **Calculation of $Z_j^1(\tau)$:**

$Z_j^1(\tau)$ corresponds to the error coming, at the following instants, from the synapses connecting the neurons:

$$Z_j^1(\tau) = \sum_{k=1}^{n} \frac{\partial C}{\partial y_k(\tau + \Delta t)} \frac{\partial y_k(\tau + \Delta t)}{\partial x_k(\tau)} \frac{\partial x_k(\tau)}{\partial y_j(\tau)}$$

$$= \sum_{k=1}^{n} \left[ Z_k(\tau + \Delta t) \cdot S_k \cdot f'(x_k(\tau)) \cdot \omega_{jk}(\tau) \right] \quad (B1)$$

- **Calculation of $Z_j^2(\tau)$ :**

$Z_j^2(\tau)$ represents the error coming directly from the output neurons. It does not exist if the neuron is hidden. Hence, for the output neurons only, there is $e_j(\tau) \neq 0$ :

- for a hidden neuron : $Z_j^2(\tau) = 0$.
- for an output neuron $Z_j^2(\tau) \neq 0$:

The calculation of the instantaneous errors $e_j(\tau)$ constitutes the coupling between the controlled system and the learning algorithm. This calculation depends on the criterion to minimize. Therefore, it is not possible to write a general formula to obtain these values because they are related to the controlled system and to the cost function:

$$Z_j^2(\tau) = \frac{\partial E}{\partial y_j(\tau)} = \frac{\partial [\sum_{k=1}^{n} e_k(\tau)]}{\partial y_j(\tau)} \Delta t = \frac{\partial e_j(\tau)}{\partial y_j(\tau)} \Delta t \quad (B2)$$

- **Calculation of $Z_j^3(\tau)$:**

$Z_j^3(\tau)$ corresponds to the error coming at the following instants by means of the internal dynamics of the neuron itself :

$$Z_j^3(\tau) = \frac{\partial C}{\partial y_j(\tau + \Delta t)} \frac{\partial y_j(\tau + \Delta t)}{\partial y_j(\tau)} = Z_j(\tau + \Delta t)(1 - S_j) \quad (B3)$$

The difference between $Z_j^1(\tau)$ and $Z_j^3(\tau)$, in the error expression coming from the $j^{th}$ neuron, comes from the terms considered in the equation (4) because $Z_j^3(\tau)$ takes into account the dynamic part expressed by the term, $(1 - S_j) y_j(t)$.

- **Calculation of $Z_j(\tau)$ :**

Finally, by gathering the equations (B.1), (B.2) and (B.3), the following equation of the backpropagated cost is obtained:

$$Z_j(\tau) = Z_j^1(\tau) + Z_j^2(\tau) + Z_j^3(\tau)$$

$$= \sum_{k=1}^{n} \left[ Z_k(\tau + \Delta t) \cdot S_k \cdot f'(x_k(\tau)) \cdot \omega_{jk}(\tau) \right] + \frac{\partial e_j(\tau)}{\partial y_j(\tau)} \cdot \Delta t + Z_j(\tau + \Delta t) \cdot (1 - S_j) \quad (B4)$$